\pdfoutput=1

\newif\ifarXiv
\arXivtrue

\newif\ifrefresh
\refreshfalse

\documentclass[runningheads]{llncs}

\usepackage[T1]{fontenc}

\usepackage[labelfont=bf]{caption}
\usepackage{subcaption}

\usepackage{booktabs}
\usepackage{bm}
\usepackage{censor}
\usepackage{dsfont}
\usepackage{etoolbox}
\usepackage{graphicx}
\usepackage{makecell}
\usepackage{multicol}
\usepackage{multirow}
\usepackage{nicefrac}
\usepackage{paralist}
\usepackage{pifont}
\usepackage{siunitx}
\usepackage{xspace}

\usepackage{tikz}
\usetikzlibrary{arrows}
\usetikzlibrary{chains}
\usetikzlibrary{positioning}
\usetikzlibrary{external}

\usepackage{pgfplots}
\usepgfplotslibrary{statistics}

\pgfplotsset{
  compat                       = 1.16,
  discard if not/.style 2 args = {
    /pgfplots/boxplot/data filter/.code={
      \edef\tempa{\thisrow{#1}}
      \edef\tempb{#2}
       \ifx\tempa\tempb
       \else
         \def\pgfmathresult{}
       \fi
    }
  }
}

\definecolor{bleu}    {RGB}{ 49,140,231}
\definecolor{cardinal}{RGB}{196, 30, 58}
\definecolor{emerald} {RGB}{ 80,200,120}

\definecolor{munich-red}   {RGB}{255, 80, 110}
\definecolor{munich-purple}{RGB}{105,  0,  95}

\pgfplotscreateplotcyclelist{miccai}{%
    munich-red,    every mark/.append style={draw opacity = 1.0, fill opacity = 0.25, fill=munich-red},    mark=*\\
    munich-purple, every mark/.append style={draw opacity = 1.0, fill opacity = 0.25, fill=munich-purple}, mark=triangle*\\
    black,         every mark/.append style={draw opacity = 1.0, fill opacity = 0.25, fill=black},         mark = square*\\
}

\pgfplotsset{
  persplot/.style = {
    axis x line*       = bottom,
    axis y line*       = left,
    enlargelimits      = false,
    unit vector ratio* = 1 1 1
    xmin               = -0.1,
    ymin               = -0.1,
    xmax               =  1.1,
    ymax               =  1.1,
    tick align         = center,
    xlabel             = {Creation},
    ylabel             = {Destruction},
    cycle list name    = {miccai},
    font               = \scriptsize,
    width              = 4cm,
    height             = 4cm,
    legend pos         = south east,
    legend style       = {%
      draw = none,
      fill = none,
    }
  },
}

\newsavebox{\featurespred}
\newsavebox{\featurestrue}

\savebox{\featurespred}{%
  \begin{tikzpicture}%
    \begin{axis}[persplot]
      \addplot+[only marks] table {Data/C_knizocytes000721_prediction_d0.txt};
      \addplot+[only marks] table {Data/C_knizocytes000721_prediction_d1.txt};
      \addplot+[only marks] table {Data/C_knizocytes000721_prediction_d2.txt};
      \addplot[domain = {-1:1}] {x};

      \legend{$d = 0$, $d = 1$, $d = 2$}
    \end{axis}
  \end{tikzpicture}%
}

\savebox{\featurestrue}{%
  \begin{tikzpicture}
    \begin{axis}[persplot]
      \addplot table {Data/C_knizocytes000721_groundtruth_d0.txt};
      \addplot table {Data/C_knizocytes000721_groundtruth_d1.txt};
      \addplot table {Data/C_knizocytes000721_groundtruth_d2.txt};

      \addplot[domain = {-1:1}] {x};
    \end{axis}
  \end{tikzpicture}
}

\usepackage{amsmath}
\usepackage{amssymb}
\usepackage{mleftright}

\newcommand{\betti}[1]      {\ensuremath{\beta_{#1}}\xspace}
\newcommand{\diagram}       {\ensuremath{\mathcal{D}}}
\newcommand{\cubicalcomplex}{\ensuremath{C}\xspace}
\newcommand{\reals}         {\ensuremath{\mathds{R}}\xspace}
\newcommand{\volume}        {\ensuremath{\mathcal{V}}\xspace}

\DeclareMathOperator{\bottleneck} {d_B}         
\DeclareMathOperator{\loss}       {\mathcal{L}}
\DeclareMathOperator{\Persistence}{Pers}        
\DeclareMathOperator{\persistence}{pers}        
\DeclareMathOperator{\wasserstein}{W}

\newcommand{\geomloss}{\loss_{\text{G}}}
\newcommand{\topoloss}{\loss_{\text{T}}}

\usepackage{hyperref}
\usepackage[nameinlink,capitalize]{cleveref}

\hypersetup{%
  colorlinks = true,
  urlcolor   = blue,
  linkcolor  = blue,
  citecolor  = blue,
}

\ifrefresh
\fi

\newcommand{\shapr}{\texttt{SHAPR}\xspace}

\begin{document}

\title{%
  Capturing Shape Information with Multi-Scale Topological Loss Terms for 3D Reconstruction
}

\titlerunning{Capturing Shape Information with Multi-Scale Topological Loss Terms}

\author{Dominik J.\,E.\ Waibel\inst{1,3}\orcidID{0000-0002-3753-8598} \and
Scott Atwell\inst{2}\orcidID{0000-0001-9943-3524}
\and
Matthias Meier\inst{2}\orcidID{0000-0001-9455-4538}\and
Carsten Marr\inst{1}\orcidID{0000-0003-2154-4552} \and
Bastian Rieck\inst{1,2,3}\orcidID{0000-0003-4335-0302}%
}%

\authorrunning{D.\,J.\,E.\ Waibel et al.}
%
\institute{Institute of AI for Health, Helmholtz Munich -- German Research Centre for Environmental Health, Neuherberg, Germany \and
Helmholtz Pioneer Campus, Helmholtz Munich-- German Research Centre for Environmental Health, Neuherberg, Germany \and 
Technical University of Munich, Munich, Germany
\email{bastian.rieck@helmholtz-muenchen.de}}

\maketitle

\begin{abstract}
Reconstructing 3D objects from 2D images is both challenging for our
brains and machine learning algorithms. To support this spatial
reasoning task, contextual information about the overall shape of an
object is critical. However, such information is not captured by
established loss terms (e.g.\ Dice loss). 
We propose to complement geometrical shape information by including
multi-scale topological features, such as connected components, cycles,
and voids, in the reconstruction loss. Our method uses cubical complexes
to calculate topological features of 3D volume data and employs an
optimal transport distance to guide the reconstruction process. This
topology-aware loss is fully differentiable, computationally efficient,
and can be added to any neural network.
We demonstrate the utility of our loss by incorporating it into \shapr,
a model for predicting the 3D cell shape of individual cells based on 2D
microscopy images. Using a hybrid loss that leverages both geometrical
and topological information of single objects to assess their shape, we
find that topological information substantially improves the quality of
reconstructions, thus highlighting its ability to extract more relevant
features from image datasets.
\end{abstract}

\section{Introduction}

Segmentation and reconstruction are common tasks when dealing with
imaging data. Especially in the biomedical domain, segmentation accuracy
can have a substantial impact on complex downstream tasks, such as
a patient's diagnosis and treatment.
3D segmentation is a complex task in itself, requiring the assessment and
labelling of each voxel in a volume, which, in turn, necessitates
a high-level understanding of the object and its context. However,
complexity rapidly increases when attempting to reconstruct a 3D object
from a 2D projection since 3D images may often be difficult to obtain.
This constitutes an inverse problem with intrinsically ambiguous
solutions: each 2D image permits numerous 3D reconstructions, similar to
how a shadow alone does not necessarily permit conclusions to be drawn
about the corresponding shape.
When addressed using machine learning, the solution of such inverse
problems can be facilitated by imbuing a model with additional inductive
biases about the structural properties of objects.
Existing models supply such inductive biases to the reconstruction task
mainly via geometry-based objective functions, thus learning a likelihood
function $f\colon\volume\to\reals$, where $f(x)$ for $x \in \volume$ denotes
the likelihood that a voxel~$x$ of the input volume~$\volume$
is part of the ground truth shape~\cite{9356353}.
Loss functions to learn~$f$ are evaluated on a per-voxel basis,
assessing the differences between the original volume and the predicted
volume in terms of overlapping labels.
Commonly-used loss functions include binary cross entropy~(BCE), Dice
loss, and mean squared error~(MSE).
Despite their expressive power, these loss terms do not capture
structural shape properties of the volumes.

\paragraph{Our contributions.}
Topological features, i.e.\
features that characterise data primarily in terms of
\emph{connectivity}, have recently started to emerge as a powerful
paradigm for complementing existing machine learning
methods~\cite{Hensel21}. They are capable of capturing shape information
of objects at multiple scales and along multiple dimensions. In this
paper, we leverage such features and integrate them into a novel differentiable `topology-aware' loss term
$\topoloss$ that can be used to regularise the shape reconstruction
process. Our loss term handles arbitrary shapes, can be
computed efficiently, and may be integrated into general deep learning
models.
We demonstrate the utility of~$\topoloss$ by combining it with
\shapr~\cite{Waibel2021-or}, a framework for predicting individual cell
shapes from 2D microscopy images.
The new hybrid variant of \shapr, making use of both geometry-based and
topology-based objective functions, results in improved reconstruction
performance along multiple metrics.

\section{Related Work}

Several deep learning approaches for predicting 3D shapes of single
objects from 2D information already exist; we aim to give a brief
overview.
Previous work includes predicting natural objects such as air planes, cars, and
furniture from photographs, creating either meshes~\cite{Gkioxari2019-ba,Wang2018-qg},
voxel volumes~\cite{Choy2016-fa}, or point clouds~\cite{Fan2017-uf}.
A challenging biomedical task, due to the occurrence of imaging noise,
is tackled by Waibel et al.~\cite{Waibel2021-or}, whose \shapr model
predicts the shape and morphology of individual mammalian cells from 2D microscopy images.
Given the multi-scale nature of microscopy images, \shapr is an ideal
use case to analyse the impact of employing additional topology-based
loss terms for these reconstruction tasks.

Such loss terms constitute a facet of the emerging field of
\emph{topological machine learning} and \emph{persistent homology}, its
flagship algorithm~(see \cref{sec:Topology} for an introduction).
Previous studies have shown great promise in using topological losses
for image segmentation tasks or their evaluation~\cite{Shit21a}.
In contrast to our loss, existing work relies on prior knowledge about
`expected' topological features~\cite{Clough20a}, or enforces
a pre-defined set of topological features based on comparing
segmentations~\cite{Hu19a,Hu21a}.

\section{Our Method: A Topology-Aware Loss}

\begin{figure}[tbp]
  \centering
  \ifrefresh%
    \begin{tikzpicture}[start chain = main going right, node distance = 6.5mm]
      \tikzset{%
          block/.style = {%
            on chain,
            align          = center,
            inner sep      = 2.5pt,
            minimum height = 1.75cm,
            minimum width  = 1.75cm,
            draw           = black,
          },
          exchangeable/.style = {%
            dotted,
          },
          every label/.style    = {%
            font = \scriptsize,
          },
          >=stealth',
      }
      \node[block, label = above:{2D Input}] (Input) {%
        \includegraphics[width=1cm]{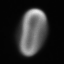}%
      };

      \node[block, exchangeable, label = above:{Model}] (ML) {%
        \includegraphics[height=1cm]{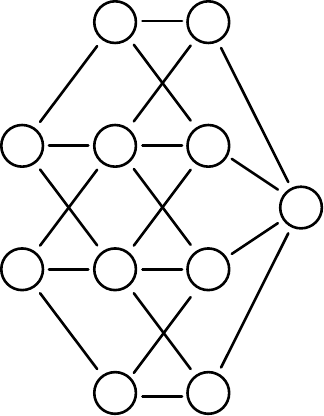}%
      };

      \node[block, label = above:{3D Prediction}] (Prediction) {%
        \includegraphics[height=1cm]{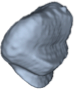}%
      };

      \node[block, exchangeable, label = above:{Geometrical loss}] (GLoss) {%
        $\geomloss$%
      };

      \node[block, label = above:{3D Ground Truth}] (Truth) {%
        \includegraphics[height=1.5cm]{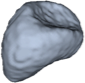}%
      };

      \node[
        block,
        label = below:{Topological features},
        below = 5mm of Prediction
      ] (PFeatures) {%
        \resizebox{1.65cm}{!}{%
          \usebox{\featurespred}%
        }%
      };

      \node[%
        block,
        label      = below:{\bfseries Topological loss},
        below      = 5mm of GLoss, 
        thick,
      ] (TLoss) {%
        $\topoloss$%
      };

      \node[
        block,
        label     = below:{Topological features},
        below     = 5mm of Truth,
        inner sep = 0pt,
      ] (TFeatures) {%
        \resizebox{1.65cm}{!}{%
          \usebox{\featurestrue}%
        }%
      };

      \draw[->] (Input)      edge (ML);
      \draw[->] (ML)         edge (Prediction);
      \draw[->] (Prediction) edge (GLoss);
      \draw[->] (Truth)      edge (GLoss);
      \draw[->] (Prediction) edge (PFeatures);
      \draw[->] (Truth)      edge (TFeatures);
      \draw[->] (PFeatures)  edge (TLoss);
      \draw[->] (TFeatures)  edge (TLoss);
      \path (TLoss) -- node[midway]{\scalebox{1.25}{$\oplus$}} (GLoss);
    \end{tikzpicture}
  \else
    \includegraphics[width=\linewidth]{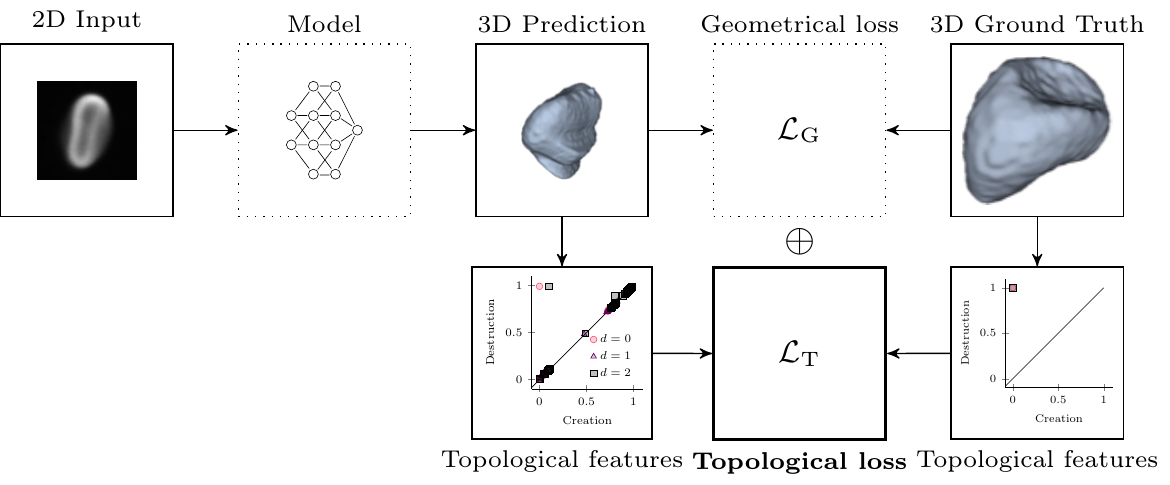}
  \fi
  \caption{%
    Given a predicted object and a 3D ground truth object, we calculate
    topological features using cubical persistent homology, obtaining
    a set of persistence diagrams.
    Each point in a persistence diagram denotes the creation and
    destruction of a $d$-dimensional topological feature of the given
    object.
    We compare these diagrams using $\topoloss$, our novel
    topology-based loss, combining it with geometrical loss terms such
    as binary cross entropy~(BCE). Dotted components can be swapped out.
  }
  \label{fig:Overview}
\end{figure}
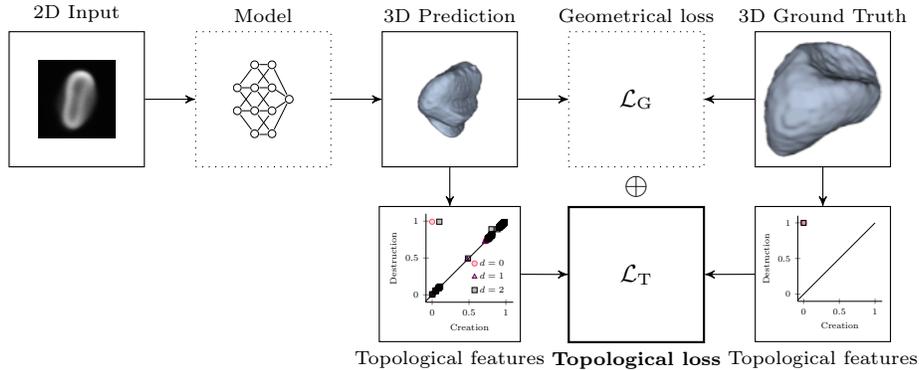

We propose a topology-aware loss term based on concepts from topological
machine learning and optimal transport. The loss term works on the level
of individual volumes, leveraging a valid metric between topological
descriptors, while remaining efficiently computable. Owing to its
generic nature, the loss can be easily integrated into existing architectures;
see \cref{fig:Overview} for an overview.

\subsection{Assessing the Topology of Volumes}\label{sec:Topology}

Given a volume~\volume, i.e.\ a \mbox{$d$-dimensional} tensor of shape
$n_1 \times n_2 \times \dots \times n_d$, we represent it as
a \emph{cubical complex}~\cubicalcomplex. A cubical complex contains
individual voxels of~\volume as
vertices~\includegraphics[height=0.60em]{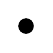},
along with connectivity information about their neighbourhoods via 
edges~\includegraphics[height=0.60em]{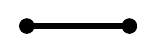},
squares~\includegraphics[height=1em]{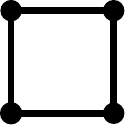},
and their higher-dimensional
counterparts.\footnote{%
  Expert readers may recognise that cubical complexes are related to
  meshes and simplicial complexes but use squares instead of triangles
  as their building blocks.
}
Cubical complexes provide a fundamental way to represent
volume data and have proven their utility in previous
work~\cite{Rieck20,Wagner12}.
Topological features of different dimensions are well-studied,
comprising connected components~(0D), cycles~(1D), and voids~(2D), for
instance.
The number of \mbox{$k$-dimensional} topological features is also
referred to as the $k$th \emph{Betti number}~\betti{k}
of~\cubicalcomplex.
While previous work has shown the efficacy of employing Betti
numbers as a topological prior for image segmentation
tasks~\cite{Clough20a,Hu19a}, the reconstruction tasks we are considering in this paper require a multi-scale perspective
that cannot be provided by Betti numbers, which are mere feature counts.
We therefore make use of \emph{persistent
homology}, a technique for calculating multi-scale topological
features~\cite{Edelsbrunner02}.
This technique is particularly appropriate in our setting: our model
essentially learns a likelihood function~$f\colon\volume\to\reals$.
To each voxel $x \in
\volume$, the function~$f$ assigns the likelihood of~$x$ being part of
an object's shape.
For a likelihood threshold~$\tau\in\reals$, we obtain a cubical
complex $\cubicalcomplex^{(\tau)} := \{x \in \volume \mid f(x) \geq \tau\}$
and, consequently, a different set of topological features.
Since volumes are finite, their topology only
changes at a finite number of thresholds $\tau_1 \geq \dotsc \geq \tau_m$,
and we obtain a sequence of nested cubical complexes
  $\emptyset \subseteq C^{(\tau_1)} \subseteq C^{(\tau_2)} \subseteq \dots
  \subseteq C^{(\tau_m)} = \volume$,
known as a \emph{superlevel set filtration}. Persistent homology
tracks topological features across all complexes in this filtration,
representing each feature as a tuple $(\tau_i, \tau_j)$, with
$\tau_i \geq \tau_j$, indicating the cubical complex in which a feature
was being `created' and `destroyed,' respectively. The tuples
of \mbox{$k$-dimensional} features, with $0 \leq k \leq d$,
are stored in the $k$th
\emph{persistence diagram}~$\diagram_f^{(k)}$ of the data
set.\footnote{%
  We use the subscript~$f$ to indicate the corresponding likelihood
  function; we will drop this for notational convenience when discussing
  general properties.
}
Persistence diagrams thus form a multi-scale shape
descriptor of all topological features of a dataset. Despite
the apparent complexity of filtrations, persistent homology of
cubical complexes can be calculated efficiently in practice~\cite{Wagner12}.

\paragraph{Structure of persistence diagrams.} 
Persistent homology provides information
beyond Betti numbers: instead of enforcing a choice of
threshold~$\tau$ for the likelihood function, which would result in a fixed set of Betti numbers, persistence diagrams
encode all thresholds at the same time, thus capturing additional geometrical details
about data.
Given a tuple $(\tau_i, \tau_j)$ in a persistence diagram, its
\emph{persistence} is defined as $\persistence\mleft(\tau_i,
\tau_j\mright) := \mleft|\tau_j - \tau_i\mright|$. Persistence indicates
the `scale' over which a topological feature occurs, with large values
typically assumed to correspond to more stable features. 
The sum of all persistence values is known as the \emph{\mbox{degree-$p$} total
persistence}, i.e.\ $\Persistence_p(\diagram_f) := \sum_{(\tau_i,
\tau_j) \in \diagram_f} \mleft|\persistence(\tau_i, \tau_j)^p\mright|$.
It constitutes a stable summary statistic of topological
activity~\cite{Cohen-Steiner10}.

\paragraph{Comparing persistence diagrams.}
Persistence diagrams can be endowed with a metric
by using optimal transport. Given two diagrams
$\diagram$ and $\diagram'$ containing features of the same dimensionality,
their $p$th \emph{Wasserstein distance} is defined as 
\begin{equation}
  \wasserstein_p\mleft(\diagram, \diagram'\mright) := \mleft( \inf_{\eta\colon \diagram \to \diagram'}\sum_{x\in\diagram}\|x-\eta(x)\|_\infty^p \mright)^{\frac{1}{p}},
  \label{eq:Wasserstein}
\end{equation}
where $\eta(\cdot)$ denotes a bijection.
Since $\diagram$ and $\diagram'$ generally have different
cardinalities, we consider them to contain an infinite number of points
of the form $(\tau, \tau)$, i.e.\ tuples of zero persistence.
A suitable~$\eta(\cdot)$ can thus always be found. Solving
\cref{eq:Wasserstein} is practically feasible using  modern
optimal transport algorithms~\cite{Flamary21a}.

\paragraph{Stability.}
%
A core property of persistence diagrams is their stability to noise.
While different notions of stability exist for persistent
homology~\cite{Cohen-Steiner10}, a recent theorem~\cite{Skraba21a}
states that the Wasserstein distance between persistence diagrams of
functions~$f, f'\colon\volume \to \reals$ is bounded by their $p$-norm,
i.e.\ 
\begin{equation}
  \wasserstein_p\mleft(\diagram_f^{(k)}, \diagram_{f'}^{(k)}\mright)
  \leq C \mleft\| f - f'\mright\|_p \text{ for $0 \leq k \leq d$},
  \label{eq:Stability}
\end{equation}
with~$C \in \reals_{> 0}$ being a constant that depends on the dimensionality
of~$\volume$. \cref{eq:Stability} implies that gradients obtained from
the Wasserstein distance and other topology-based summaries will remain
bounded; we will also use it to accelerate
topological feature calculations in practice.

\paragraph{Differentiability.}

Despite the discrete nature of topological features, persistent homology
permits the calculation of gradients with respect to parameters of the likelihood function~$f$, thus enabling the use of automatic
differentiation schemes~\cite{Hofer20a,Moor20a,Poulenard18}.
A seminal work by Carri\`ere et al.~\cite{Carriere21a} proved that
optimisation algorithms converge for a wide class of persistence-based functions, thus
opening the door towards general topology-based optimisation schemes.

\subsection{Loss Term Construction}

Given a true likelihood function~$f$ and a predicted likelihood
function~$f'$, our novel generic topology-aware loss term takes the form 
\begin{equation}
  \topoloss\mleft(f, f', p\mright) := \sum_{i=0}^{d} \wasserstein_p\mleft(\diagram_f^{(i)},
  \diagram_{f'}^{(i)}\mright) + \Persistence\mleft(\diagram_{f'}^{(i)}\mright).
  \label{eq:Loss}
\end{equation}
The first part of \cref{eq:Loss} incentivises the model to reduce the
distance between $f$ and $f'$ with respect to their topological shape
information.
The second part incentivises the model to reduce overall topological
activity, thus decreasing the noise in the reconstruction. This can be
considered as the topological equivalent of reducing the \emph{total
variation} of a function~\cite{Rieck16a}.
Given a task-specific geometrical loss term $\geomloss$,\footnote{%
We dropped all hyperparameters of the loss term
for notational clarity.
} such as
a Dice loss, we obtain a combined loss term as $\loss :=
\geomloss+ \lambda \topoloss$, where $\lambda \in \reals_{> 0}$ controls
the impact of the topology-based part.
We will use $p = 2$ since \cref{eq:Stability} relates $\topoloss$ to the
Euclidean distance in this case.

\paragraph{Calculations in practice.}
%
To speed up the calculation of our loss term, we utilise the stability
theorem of persistent homology and downsample each volume to $M \times
M \times M$ voxels using trilinear interpolation.
We provide a theoretical and empirical analysis of the errors introduced
by downsampling in the Supplementary Materials. In our experiments, we
set $M = 16$, which is sufficiently small to have no negative impact
on computational performance while at the same time resulting 
in empirical errors $\leq 0.1$~(measured using \cref{eq:Stability} for $p = 2$).

\section{Experiments}

We provide a brief overview of \shapr before discussing the experimental
setup, datasets, and results.
\shapr is a deep learning method to predict 3D shapes of single cells
from 2D microscopy images~\cite{Waibel2021-or}. Given a 2D fluorescent
image of a single cell and a corresponding segmentation mask, \shapr
predicts the 3D shape of this cell.
The authors suggest to train \shapr with a combination of Dice and
BCE loss, with an additional adversarial training step
to improve the predictions of the model.
We re-implemented \shapr using
\texttt{PyTorch}~\cite{PyTorch} to ensure fair comparisons and the seamless integration of our novel
loss function~$\topoloss$, employing `Weights~\&~Biases'
for tracking experiments~\cite{wandb}.
Our code and
reports are publicly available.\footnote{
  See {\url{https://github.com/marrlab/SHAPR_torch}}.
}

\paragraph{Data.}
For our experiments, we use the two datasets published with the original \shapr
manuscript~\cite{Waibel2021-or}.
The first dataset comprises $825$ red blood cells, imaged in 3D with
a confocal microscope~\cite{Simionato21a}.
Each cell is assigned to one of nine pre-defined classes:
sphero-,
stomato-,
disco-,
echino-,
kerato-,
knizo-, and
acanthocytes, as well as
cell clusters and
multilobates.
The second dataset contains $887$ nuclei of human-induced pluripotent
stem cells~(iPSCs), counterstained and imaged in 3D with a confocal
microscope. Cells were manually segmented to create ground truth objects. Both
datasets include 3D volumes of size $64 \times 64 \times 64$, 2D
segmentation masks of size $64 \times 64$, and fluorescent images of
size $64 \times 64$.

\subsection{Training and Evaluation}

We trained our implementation of \shapr for a maximum of $100$ epochs, using early
stopping with a patience of $15$ epochs, based on the validation loss.
For each run, we trained five \shapr models in a round-robin fashion,
partitioning the dataset into five folds with a 60\%/20\%/20\%
train/validation/test split, making sure that each 2D input image
appears once in the test set.
To compare the performance of \shapr with and without $\topoloss$, we
used the same hyperparameters for all experiments~(initial learning rate
of \num{1e-3}, $\beta_1 = 0.9$, and $\beta_2 =0.999$ for the ADAM
optimiser). We optimised $\lambda \in \{\num{1e-3}, \num{1e-2}, \dots,
\num{1e2}\}$, the regularisation strength
parameter for $\topoloss$, on an independent dataset, resulting in
$\lambda = 0.1$ for all experiments. We also found that evaluating
\cref{eq:Loss} for each dimension individually leads to superior
performance; we thus only calculate \cref{eq:Loss} for $i = 2$.
Finally, for the training phase, we augmented the data with random
horizontal or vertical flipping and $90^\circ$ rotations with a 33\%
chance for each augmentation to be applied for a sample. The goal of
these augmentations is to increase data variability and prevent
overfitting.

Following Waibel et al.~\cite{Waibel2021-or}, we
evaluated the performance of \shapr by calculating
\begin{inparaenum}[(i)]
  \item the \emph{intersection over union}~(IoU) error,
  \item the relative volume error,
  \item the relative surface area error, and
  \item the relative surface roughness error
\end{inparaenum}
with respect to the ground truth data,
applying Otsu's method~\cite{Otsu79} for thresholding predicted shapes.
We calculate the volume by counting non-zero voxels, the surface area as all voxels on the surface of an object, and the surface roughness as the difference between the surface area of the object and the surface area of the same object after smoothing it with a 3D Gaussian~\cite{Waibel2021-or}. 

\begin{figure}[tbp]
\centering
  \subcaptionbox{\label{sfig:Examples}}{%
    \includegraphics[height=6cm]{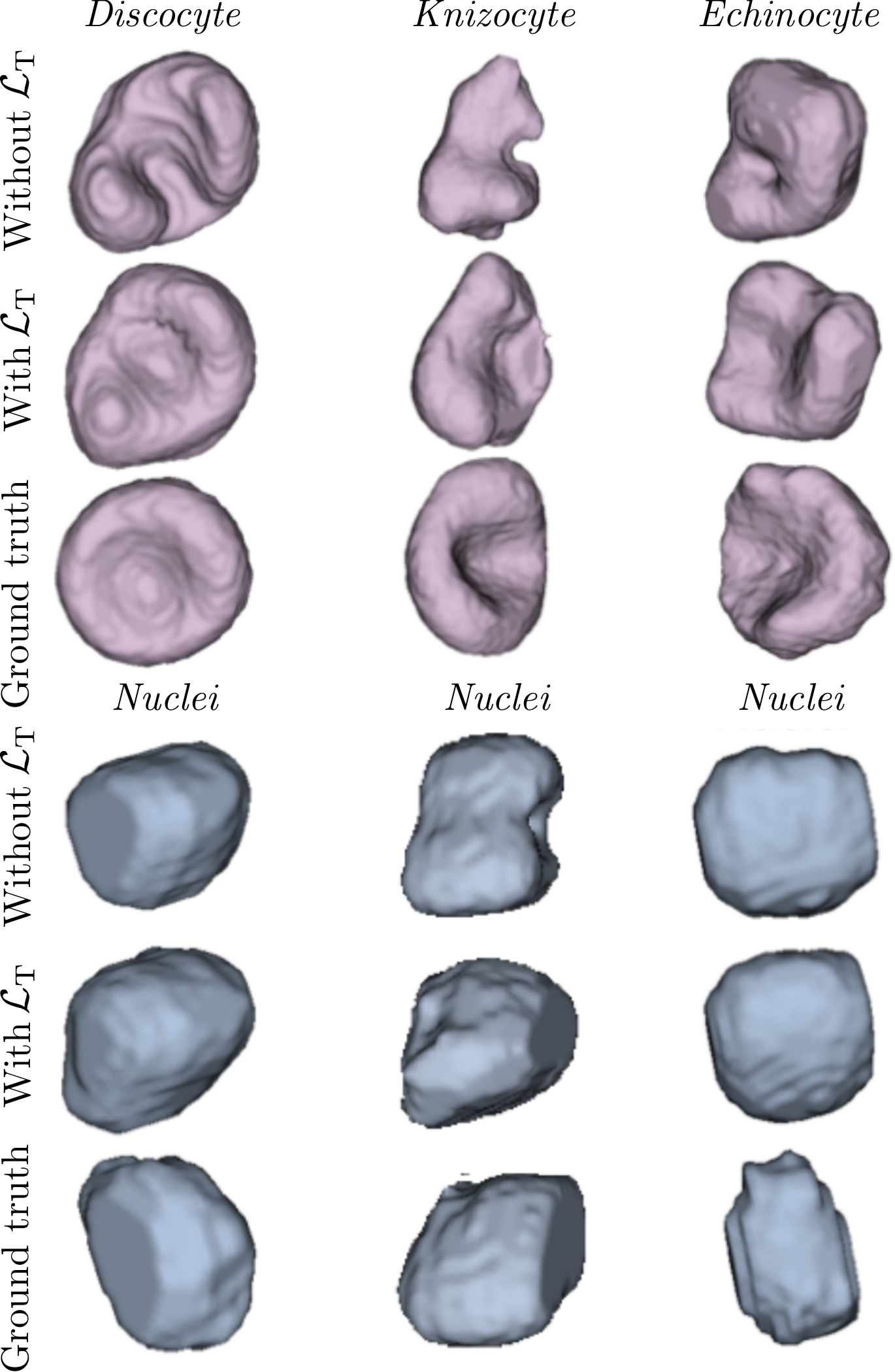}%
  }%
  \subcaptionbox{\label{sfig:Swarms}}{%
    \begin{tabular}{@{}l@{}}
    \includegraphics[width=8cm]{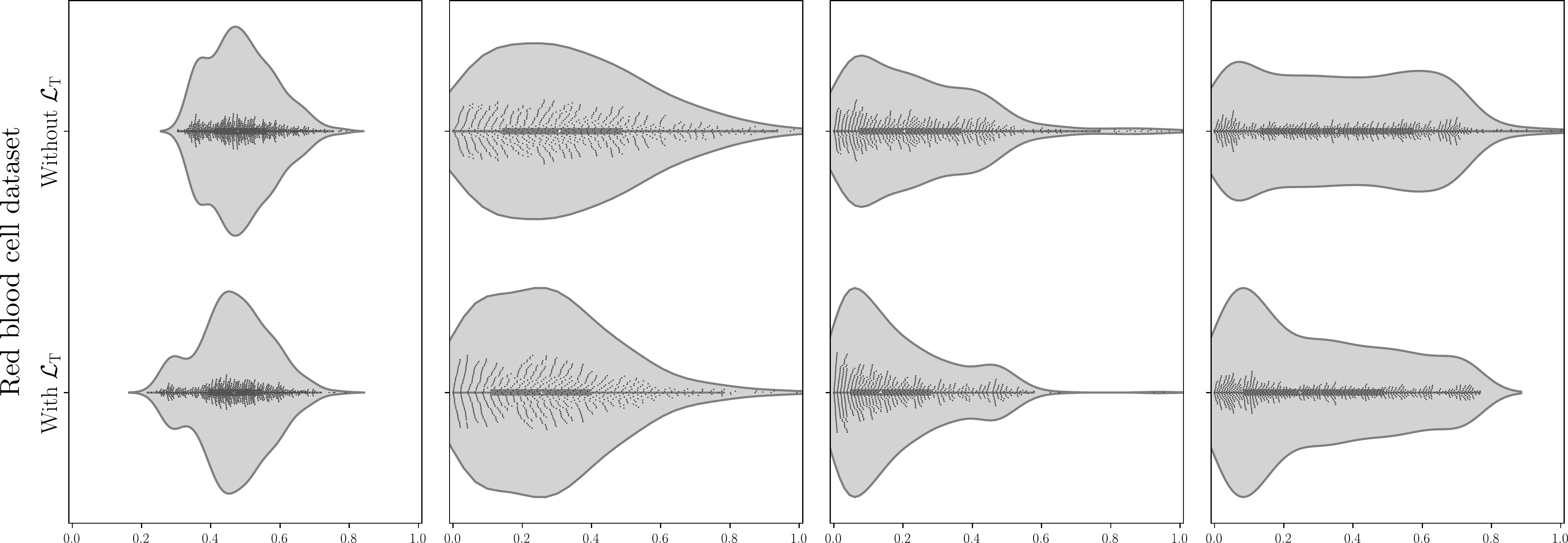}\\
    \includegraphics[width=8cm]{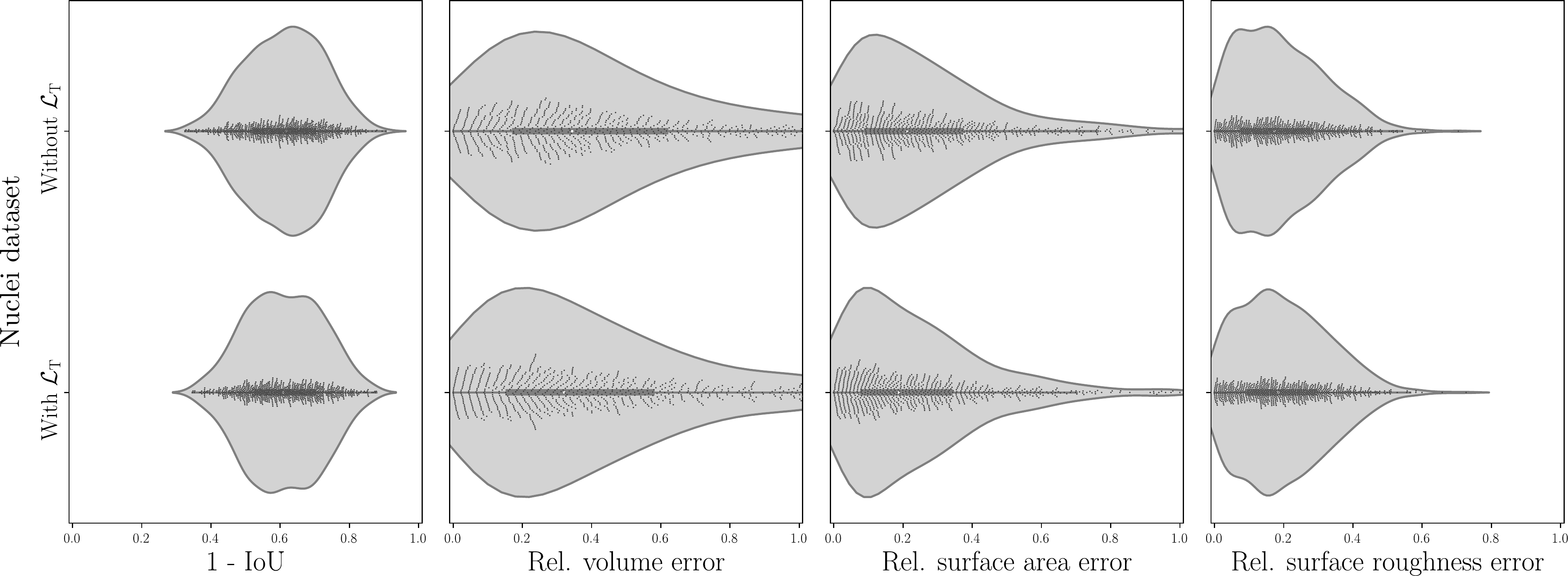}%
    \end{tabular}%
  }
  \caption{%
    \subref{sfig:Examples}~Examples of predictions without~(top row) and
    with~(middle row) $\topoloss$, our topological loss term. The third
    row shows ground truth images.
    \subref{sfig:Swarms}~$\topoloss$ improves predictions in relevant
    metrics, such as the IoU error, the relative volume error, relative
    surface area error, and relative surface roughness error.
  }
  \label{fig:Results}
\end{figure}

\subsection{Results}

To evaluate the benefits of our topology-aware loss, we perform the same
experiment twice: first, using a joint BCE and Dice
loss~\cite{Waibel2021-or}, followed by adding $\lambda \topoloss$.
Without $\topoloss$, we achieve results comparable to the original
publication~(IoU, red blood cell data: $0.63\pm0.12\%$; IoU, nuclei
data: $0.46\pm0.16\%$); minor deviations arise from stochasticity and
implementation differences between \texttt{PyTorch} and
\texttt{Tensorflow}.
We observe superior performance in the majority of metrics for both
datasets when adding $\topoloss$ to the model~(see \cref{fig:Results}
and \cref{tab:Results}); the performance gains by $\topoloss$ are
statistically significant in all but one case.
Notably, we find that $\topoloss$ increases \shapr's predictive
performance unevenly across the classes of the red blood cell
dataset~(see \cref{sfig:Examples}). For spherocytes~(round cells),
only small changes in IoU error, relative volume error, relative
surface area error, and relative surface roughness error~(7\% decrease, 2\% decrease, and 3\% decrease,
respectively) occur, whereas for echinocytes~(cells with a spiky
surface), we obtain a 27\% decrease in IoU error and 7\% decrease in
relative volume error.
Finally, for discocytes~(bi-concave cells) and
stomatocytes, we obtain a 2\% decrease in IoU error, a 3\% decrease in
volume error,  a 11\% decrease in surface area error, and a  25\% decrease in surface roughness error upon adding
$\topoloss$.

\begin{table}[btp]
  \centering
  \caption{
    Median, mean~($\mu$) and standard deviation~($\sigma$) of several
    relative error measures for two datasets~(lower values are better;
    winner shown in \textbf{bold}).
    The $\topoloss$ column indicates whether our new loss term was
    active. We also show the $p$-value of a paired Wilcoxon
    signed-rank test between error distributions.
  }
  \label{tab:Results}
  \sisetup{
    detect-all           = true,
    table-format         = 1.2(3),
    round-mode           = places,
    round-precision      = 2,
    separate-uncertainty = true,
    table-text-alignment = center,
  }
  \newcommand{\yes}{\ding{51}}
  \newcommand{\no} {\ding{55}}

  \robustify\bfseries
  \renewrobustcmd{\bfseries}{\fontseries{b}\selectfont}
  \renewrobustcmd{\boldmath}{}
  \let\b\bfseries

  \begin{tabular}{llS[table-format=1.2]SS[round-precision=1]S[table-format=1.3]SS[round-precision=1]}
    \toprule
    & & \multicolumn{3}{c}{Red blood cell~($n = 825$)} & \multicolumn{3}{c}{Nuclei~($n = 887$)}\\
    \midrule
    Relative error & $\topoloss$ & {Median} & {$\mu\pm\sigma$} & {$p$} & {Median} & {$\mu\pm\sigma$} & {$p$}\\
    \midrule
    \multirow{2}{*}{$1-\mathrm{IoU}$}             & \no  & 0.48 & 0.49 \pm 0.09 & {\multirow{2}{*}{\num{1.1E-19}}} &  0.62&  0.62\pm 0.11 & {\multirow{2}{*}{\num{0.491}}}\\
                                       & \yes & \b0.47 & \b0.47 \pm 0.10 &                                  &\b0.61&\b0.61\pm 0.11 &\\
    \midrule
    \multirow{2}{*}{Volume}            & \no  & 0.31 & 0.35 \pm 0.31 & {\multirow{2}{*}{\num{1.2E-47}}} &  0.34 &  0.48 \pm 0.47 & {\multirow{2}{*}{\num{4.6E-14}}}\\
                                       & \yes & \b0.26 & \b0.29 \pm 0.27 &                                  &\b0.32 &\b0.43 \pm 0.42 &\\ 
    \midrule
    \multirow{2}{*}{Surface area}      & \no  & 0.20 & 0.24 \pm 0.20 & {\multirow{2}{*}{\num{39.5E-13}}}          &  0.21 &  0.27 \pm 0.25 & {\multirow{2}{*}{\num{1.7E-8}}}\\
                                       & \yes & \b0.14 & \b0.18 \pm 0.16 &                                  &\b0.18 &\b0.25 \pm 0.24 &\\
    \midrule
    \multirow{2}{*}{Surface roughness} & \no  & 0.35 & 0.36 \pm 0.24 & {\multirow{2}{*}{\num{9.1E-4}}}          &\b0.17 &\b0.18 \pm 0.12 & {\multirow{2}{*}{\num{1.5E-6}}}\\
                                       & \yes & \b0.24 & \b0.29 \pm 0.22 &                                  &  0.18 &  0.19 \pm 0.13 &\\ 
    \bottomrule
  \end{tabular}
\end{table}

\section{Discussion}

We propose a novel topology-aware loss term~$\topoloss$ that can be
integrated into existing deep learning models.
Our loss term is not restricted to specific types of shapes and may
be applied to reconstruction and segmentation tasks.
We demonstrate its efficacy in the reconstruction of 3D shapes from 2D
microscopy images, where the results of \shapr are statistically
significantly improved in relevant metrics whenever $\topoloss$
is jointly optimised together with geometrical loss terms.
Notably, $\topoloss$ does not
optimise classical segmentation/reconstruction metrics, serving instead
as an inductive bias for incorporating multi-scale information on
topological features.
$\topoloss$ is computationally efficient and can be adapted to
different scenarios by incorporating topological features of a specific
dimension.
Since our experiments indicate that the calculation of \cref{eq:Loss}
for a single dimension is sufficient to achieve improved reconstruction
results in practice, we will leverage topological duality/symmetry
theorems~\cite{Cohen-Steiner09} in future work to improve computational
efficiency and obtain smaller cubical complexes.

Our analysis of predictive performance across classes of the red blood
cell dataset~(see \cref{tab:Results}) leads to the assumption that the
topological loss term shows the largest reconstruction performance
increases on shapes that have complex morphological features, such as
echinocytes or bi-concavely shaped cells~(discocytes and stomatocytes).
This implies that future extensions of the method should
incorporate additional geometrical descriptors into the filtration
calculation, making use of recent advances in capturing the topology of
multivariate shape descriptors~\cite{Lesnick15}.

\subsubsection{Acknowledgements.} 
We thank Lorenz Lamm, Melanie Schulz, Kalyan Varma Nadimpalli, and
Sophia Wagner for their valuable feedback to this manuscript. The
authors also are indebted to Teresa Heiss for discussions on the
topological changes induced by downsampling volume data.

\subsubsection{Funding.}
Carsten Marr received funding from the European Research Council~(ERC)
under the European Union's Horizon 2020 Research and Innovation
Programme~(Grant Agreement~866411).

\subsubsection{Author contributions.} 
DW and BR implemented code and conducted experiments. DW, BR, and CM
wrote the manuscript. DW created figures and BR the main portrayal of
results. SA and MM provided the 3D nuclei dataset. BR supervised the study. All
authors have read and approved the manuscript.


\ifarXiv
  \appendix
\section{Discussion: The Effects of Interpolation}

As outlined in the main paper, we downsample our volumes from $64^3$
voxels to $16^3$ voxels using \emph{trilinear interpolation}. To assess
the impact this has on the calculation of persistent homology, we can
make use of recent work on the topological impact of changing image
resolutions~\cite{Heiss21a}. Following Heiss et al.~\cite{Heiss21a}, we
treat downsampling as a general function that transforms a volume of
side length $r_1 \in \mathds{N}$ into a volume of side length $r_2 \in
\mathds{N}$.
Assuming that $r_1 > r_2 > 0$ and $r_2 \mid r_1$, we refer to $a :=
\nicefrac{r_1}{r_2}$ as the \emph{compression factor}. In $d$
dimensions, every voxel of size~$r_2$ contains~$a^d$ voxels of
size~$r_1$, making it possible to compare the two volumes by repeating
the values of the smaller volume.

While the bounds given by Heiss et al.~\cite{Heiss21a} make use of the
\emph{bottleneck distance} instead of the Wasserstein distance, we can
still use the equivalence of norms to obtain bounds. To this end, we
first observe that the likelihood function~$f$ that we use in the paper
is Lipschitz continuous with Lipschitz constant~$L = 1$. This is
a direct consequence of the image of the function being restricted to
$[0, 1]$. Corollary III.4 of Heiss et al.~\cite{Heiss21a} now states the
following:\footnote{%
  We use a slightly different formulation that is
  aligned with the notation of our main paper. Moreover, we consider
  $r_1$ to refer to the larger side length, as outlined above.
}
\begin{corollary}
  If $f\colon \reals^d \to \reals$ is Lipschitz continuous with
  Lipschitz constant~$L$, then $\bottleneck(\diagram_{f_{r_1}},
  \diagram_{f_{r_2}}) \leq L r_1 \sqrt{d}$, where $\bottleneck(\cdot,
  \cdot)$ refers to the bottleneck distance between two persistence
  diagrams, and $\diagram_{f_{r_1}}, \diagram_{f_{r_2}}$ denote the
  persistence diagrams of the likelihood functions for volumes with side
  lengths $r_1$ and $r_2$, respectively.
\end{corollary}
This corollary makes use of the fact that $\|f_{r_1} - f_{r_2}\|_\infty
\leq L r_1 \sqrt{d}$, such that the bound follows as a consequence
of the stability theorem of persistent homology~\cite{Cohen-Steiner07,Skraba21a}.
Using the equivalence of norms in finite-dimensional vector spaces, we
know that for every $x \in \reals^n$,
\begin{equation}
  \|x\|_\infty \leq \|x\|_2 \leq \sqrt{n} \|x\|_\infty.
  \label{eq:Equivalence}
\end{equation}
Treating the values of our likelihood function as a vector in $n :=
r_1^d$ dimensions, we obtain an upper bound according to \cref{eq:Equivalence} as
\begin{equation}
  \|f_{r_1} - f_{r_2}\|_2 \leq L r_1 \sqrt{d} \sqrt{r_1^{d}}.
\end{equation}
This is a worst-case bound; it does not incorporate the fact that as
$r_2$ approaches $r_1$, interpolation errors decrease. Moreover, the
bound does not incorporate the structure of the interpolation scheme;
for specific types of interpolation schemes, tighter bounds can be
derived.

\paragraph{Empirical errors.} In practice, we observe substantially
smaller errors. \cref{fig:Empirical interpolation errors} depicts
empirical interpolation errors in terms of the Wasserstein distance
between persistence diagrams arising from the original space and an
interpolated variant. The computational performance gains that we get
from using only~$16^3$ voxels~(instead of $64^3$ voxels) for the topological
calculations are shown to be accompanied by small approximation errors.

\paragraph{Code.}
Along with our model, we also publish an additional script for readers
interested in reproducing these plots. To run the analysis for $s = 8$,
for instance, you may use the following call:
\begin{scriptsize}
\begin{verbatim}
python -m scripts.analyse_interpolation -s 8 -p config/small-0D.json
\end{verbatim}
\end{scriptsize}
Please refer to our code repository\footnote{%
  See \url{https://github.com/marrlab/SHAPR_torch}.
} for more details.

\begin{figure}[tbp]
  \centering
  \pgfplotsset{%
    jitter/.style = {%
      x filter/.code ={%
        \pgfmathparse{\pgfmathresult-0.5*rnd+0.25}
      },
      fill opacity = 0.5,
      draw opacity = 0.5,
    }
  }
  \begin{tikzpicture}
    \begin{axis}[%
      axis x line*           = none,
      axis y line*           = left,
      boxplot/draw direction = y,
      width                  = 5cm,
      height                 = 6cm,
      mark size              = 0.5pt,
      tick align             = outside,
      xtick                  = {1, 2, 3},
      xticklabels            = {$32$, $16$, $8$},
      xlabel                 = {Side length of interpolated volume~($r_2$)},
      ytick                  = {0, 0.1, 0.2, 0.3, 0.4, 0.5, 0.6, 0.7, 0.8, 0.9, 1.0},
      ylabel                 = {$\wasserstein_2\mleft(\diagram_{f_{r_1}}, \diagram_{f_{r_2}}\mright)$},
    ]
      \addplot[cardinal, boxplot, thick] table [y=y, discard if not={x}{32}] {Data/Interpolation.txt};
      \addplot[%
        cardinal,
        thick,
        only marks,
        jitter,
      ] table [x expr = 1, discard if not={x}{32}] {Data/Interpolation.txt};

      \addplot[bleu,     boxplot, thick] table [y=y, discard if not={x}{16}] {Data/Interpolation.txt};

      \addplot[%
        bleu,
        thick,
        only marks,
        jitter,
      ] table [x expr = 2, discard if not={x}{16}] {Data/Interpolation.txt};

      \addplot[emerald,  boxplot, thick] table [y=y, discard if not={x}{8}] {Data/Interpolation.txt};

      \addplot[%
        emerald,
        thick,
        only marks,
        jitter,
      ] table [x expr = 3, discard if not={x}{8}] {Data/Interpolation.txt};

    \end{axis}
  \end{tikzpicture}
  \caption{%
    Empirical observation errors for a small test data set. We show that
    error with respect to the second Wasserstein distance
    $\wasserstein_2$ between the persistence diagrams generated from the
    original function~($r_1$) and a lower-resolution version with side
    length~$r_2$. The mean error for $r_2 \in \{16, 32\}$ is bounded by
    $\approx 0.1$, which we deem satisfactory for subsequent
    calculations.
  }
  \label{fig:Empirical interpolation errors}
\end{figure}
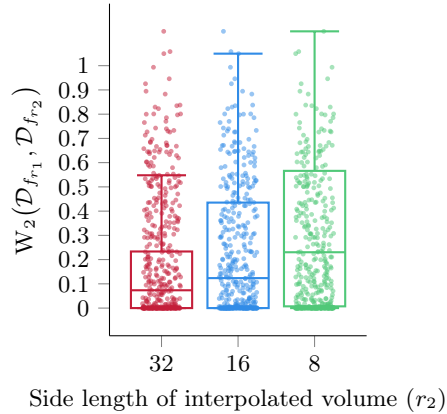

\fi

\bibliographystyle{splncs04}
\bibliography{paper2332}

\end{document}